\documentclass[runningheads]{llncs}
\usepackage{graphicx}
\usepackage{booktabs}

\usepackage{caption}
\usepackage{subcaption}
\usepackage{amsmath}
\usepackage{amssymb}
\usepackage{enumitem}

\usepackage{bm}
\usepackage{listings}
\usepackage{makecell}  
\usepackage{diagbox} 
\usepackage{mathtools} 
\usepackage{graphicx} 
\usepackage{multirow}

\usepackage{mathrsfs}

\usepackage{xcolor}
\usepackage{caption}
\usepackage{subcaption}
\usepackage{tabularx}
\usepackage{rotating}
\usepackage{verbatim}
\usepackage{xcolor,colortbl}
\usepackage{etoolbox}
\usepackage[normalem]{ulem}
\usepackage{booktabs}
\usepackage{multirow}
\usepackage{lipsum}
\usepackage{stmaryrd}
\usepackage{stackengine}
\usepackage{makecell}
\usepackage{pifont}
\usepackage{cancel}
\usepackage{adjustbox}
\usepackage{dblfloatfix}
\usepackage{bbding}
\usepackage{pifont}
\usepackage{wasysym}
\usepackage{amssymb}

\usepackage{algorithm}
\usepackage{algpseudocode}
\usepackage{amsmath}

\makeatletter
\newcommand{\car@semkitfreq}{3.92}
\newcommand{\bicycle@semkitfreq}{0.03}
\newcommand{\motorcycle@semkitfreq}{0.03}
\newcommand{\truck@semkitfreq}{0.16}
\newcommand{\othervehicle@semkitfreq}{0.20}
\newcommand{\person@semkitfreq}{0.07}
\newcommand{\bicyclist@semkitfreq}{0.07}
\newcommand{\motorcyclist@semkitfreq}{0.05}
\newcommand{\road@semkitfreq}{15.30}  %
\newcommand{\parking@semkitfreq}{1.12}
\newcommand{\sidewalk@semkitfreq}{11.13}  %
\newcommand{\otherground@semkitfreq}{0.56}
\newcommand{\building@semkitfreq}{14.1}  %
\newcommand{\fence@semkitfreq}{3.90}
\newcommand{\vegetation@semkitfreq}{39.3}  %
\newcommand{\trunk@semkitfreq}{0.51}
\newcommand{\terrain@semkitfreq}{9.17} %
\newcommand{\pole@semkitfreq}{0.29}
\newcommand{\trafficsign@semkitfreq}{0.08}
\newcommand{\semkitfreq}[1]{{\csname #1@semkitfreq\endcsname}}

\definecolor{White}{rgb}{1.,0.,1.}
\definecolor{first}{rgb}{.8,.0,.0}
\definecolor{second}{rgb}{.0,.6,.0}
\definecolor{third}{rgb}{.0,.0,.8}
\newcolumntype{g}{>{\columncolor{White}}c}

\definecolor{car.}{rgb}{0.39215686, 0.58823529, 0.96078431}
\definecolor{bicycle.}{rgb}{0.39215686, 0.90196078, 0.96078431}
\definecolor{motorcycle.}{rgb}{0.11764706, 0.23529412, 0.58823529}
\definecolor{truck.}{rgb}{0.31372549, 0.11764706, 0.70588235}
\definecolor{othervehicle.}{rgb}{0.39215686, 0.31372549, 0.98039216}
\definecolor{person.}{rgb}{1.        , 0.11764706, 0.11764706}
\definecolor{bicyclist.}{rgb}{1.        , 0.15686275, 0.78431373}
\definecolor{motorcyclist.}{rgb}{0.58823529, 0.11764706, 0.35294118}
\definecolor{road.}{rgb}{1.        , 0.        , 1.        }
\definecolor{parking.}{rgb}{1.        , 0.58823529, 1.        }
\definecolor{sidewalk.}{rgb}{0.29411765, 0.        , 0.29411765}
\definecolor{otherground.}{rgb}{0.68627451, 0.        , 0.29411765}
\definecolor{building.}{rgb}{1.        , 0.78431373, 0.        }
\definecolor{fence.}{rgb}{1.        , 0.47058824, 0.19607843}
\definecolor{vegetation.}{rgb}{0.        , 0.68627451, 0.        }
\definecolor{trunk.}{rgb}{0.52941176, 0.23529412, 0.        }
\definecolor{terrain.}{rgb}{0.58823529, 0.94117647, 0.31372549}
\definecolor{pole.}{rgb}{1.        , 0.94117647, 0.58823529}
\definecolor{trafficsign.}{rgb}{1.        , 0.        , 0.    }

\definecolor{detcolor}{gray}{.9}

\definecolor{bestcolor}{gray}{.9}

\newlength\savewidth
  
\newcolumntype{x}[1]{>{\centering\arraybackslash}p{#1pt}}
\newcolumntype{y}[1]{>{\raggedright\arraybackslash}p{#1pt}}
\newcolumntype{z}[1]{>{\raggedleft\arraybackslash}p{#1pt}}
\renewcommand{\paragraph}[1]{\vspace{1.25mm}\noindent\textbf{#1}}
\definecolor{deemph}{gray}{0.6}

\usepackage{etoolbox}
\makeatletter
\AfterEndEnvironment{algorithm}{\let\@algcomment\relax}
\AtEndEnvironment{algorithm}{\kern2pt\hrule\relax\vskip3pt\@algcomment}
\let\@algcomment\relax
\newcommand\algcomment[1]{\def\@algcomment{\footnotesize#1}}
\renewcommand\fs@ruled{\def\@fs@cfont{\bfseries}\let\@fs@capt\floatc@ruled
  \def\@fs@pre{\hrule height.8pt depth0pt \kern2pt}%
  \def\@fs@post{}%
  \def\@fs@mid{\kern2pt\hrule\kern2pt}%
  \let\@fs@iftopcapt\iftrue}
\makeatother

\definecolor{others}{rgb}{0, 0, 0}
\definecolor{barrier}{rgb}{1, 0.47058824, 0.19607843}
\definecolor{bicycle}{rgb}{1, 0.75294118, 0.79607843}
\definecolor{bus}{rgb}{1, 1, 0.0}
\definecolor{car}{rgb}{0.0, 0.58823529, 0.96078431}
\definecolor{const. veh.}{rgb}{0, 1, 1}
\definecolor{motorcycle}{rgb}{1, 0.49803922, 0}
\definecolor{pedestrian}{rgb}{1, 0, 0}
\definecolor{traffic cone}{rgb}{1, 0.94117647, 0.58823529}
\definecolor{trailer}{rgb}{0.52941176, 0.23529412, 0}
\definecolor{truck}{rgb}{0.62745098, 0.1254902, 0.94117647}

\definecolor{drive. suf.}{rgb}{1, 0, 1}
\definecolor{other flat}{rgb}{0.54509804,0.5372549,0.5372549}
\definecolor{sidewalk}{rgb}{0.29411765,0,0.29411765}
\definecolor{terrain}{rgb}{0.58823529,0.94117647,0.31372549}
\definecolor{manmade}{rgb}{0.90196078,0.90196078,0.98039216}
\definecolor{vegetation}{rgb}{0,0.68627451,0}
\definecolor{ego_vehicle}{rgb}{0,0,0}

\definecolor{turquoise}{cmyk}{0.65,0,0.1,0.1}
\definecolor{purple}{rgb}{0.65,0,0.65}
\definecolor{dark_green}{rgb}{0, 0.5, 0}
\definecolor{orange}{rgb}{0.98, 0.51, 0}
\definecolor{red}{rgb}{1.0, 0.1, 0.1}
\definecolor{blue}{rgb}{0.2, 0.2, 1.0}
\definecolor{brown}{rgb}{0.5, 0.16, 0.16}

\usepackage[accsupp]{axessibility}  

\usepackage{hyperref}

\usepackage{orcidlink}

\begin{document}
\title{SliceSemOcc: Vertical Slice–Based Multimodal 3D Semantic Occupancy Representation}
\titlerunning{Vertical Slice–Based Multimodal 3D Semantic Occupancy Representation}
%
\author{Han Huang\inst{1} 
\and Han Sun\inst{1}\textsuperscript{(\Envelope)}
\and Ningzhong Liu\inst{1}
\and Huiyu Zhou\inst{2}
\and Jiaquan Shen\inst{3}
}
\authorrunning{H.~Huang, H.~Sun et al.}
%
\institute{%
  Nanjing University of Aeronautics and Astronautics, Nanjing, China\\
  \email{huanghan\_han@163.com, sunhan@nuaa.edu.cn, lnz\_nuaa@163.com}
  \and
  University of Leicester, Leicester, UK\\
  \email{hz143@leicester.ac.uk}
  \and
  Luoyang Normal University, Luoyang, China\\
  \email{shenjiaquan\_cv@163.com}
}
\maketitle              
\begin{abstract}
Driven by autonomous driving’s demands for precise 3D perception, 3D semantic occupancy prediction has become a pivotal research topic. Unlike bird’s-eye-view (BEV) methods, which restrict scene representation to a 2D plane, occupancy prediction leverages a complete 3D voxel grid to model spatial structures in all dimensions, thereby capturing semantic variations along the vertical axis. However, most existing approaches overlook height-axis information when processing voxel features. And conventional SENet-style channel attention assigns uniform weight across all height layers, limiting their ability to emphasize features at different heights. To address these limitations, we propose SliceSemOcc, a novel vertical-slice–based multimodal framework for 3D semantic occupancy representation. Specifically, we extract voxel features along the height-axis using both global and local vertical slices. Then, a global–local fusion module adaptively reconciles fine-grained spatial details with holistic contextual information. Furthermore, we propose the SEAttention3D module, which preserves height-wise resolution through average pooling and assigns dynamic channel attention weights to each height layer. Extensive experiments on nuScenes-SurroundOcc and nuScenes-OpenOccupancy datasets verify that our method significantly enhances mean IoU, achieving especially pronounced gains on most small-object categories. Detailed ablation studies further validate the effectiveness of the proposed SliceSemOcc framework. 

\keywords{multi-modal \and 3D occupancy prediction \and vertical slice \and channel attention }
\end{abstract}
\section{Introduction}
\label{sec:intro}

LiDAR provides highly precise depth measurements and accurately captures object geometry \cite{VoxelNet,SECOND}. It maintains reliable performance under most challenging conditions, with the exception of extremely heavy rain and dense fog. The resulting 3D point clouds not only encode rich spatial structure and geometric detail, but also complement 2D image data through precise scale alignment \cite{BEVFusion,Frustum,Centerfusion,Futr3D,TransFusion,PointAugmenting}, significantly enhancing perception accuracy and robustness. In autonomous driving scenarios, these point clouds enable detailed reconstruction of spatial relationships between the ego vehicle and the surrounding obstacles, thus providing essential support for path planning and collision avoidance.

As applications such as autonomous driving \cite{Multi-view,Deep,Pv-rcnn,3dssd} and robotic navigation demand ever higher 3D perception fidelity, traditional 2D BEV detection methods \cite{BEVFormer,BEV-SAN,BEVDet4D} are constrained by their omission of height-axis information. This leads to degraded performance in multi-level structures and under complex occlusions. In contrast, 3D occupancy representations employ regular voxel grids that assign semantic labels to each voxel. By preserving hierarchical features along the height-axis, these representations support fine-grained modeling in complex scenes. Nevertheless, existing occupancy prediction models neglect height-axis hierarchies, which impedes accurate semantic characterization across different height ranges.

\begin{figure}[htbp]
\begin{center}
\includegraphics[width=1.0\linewidth]{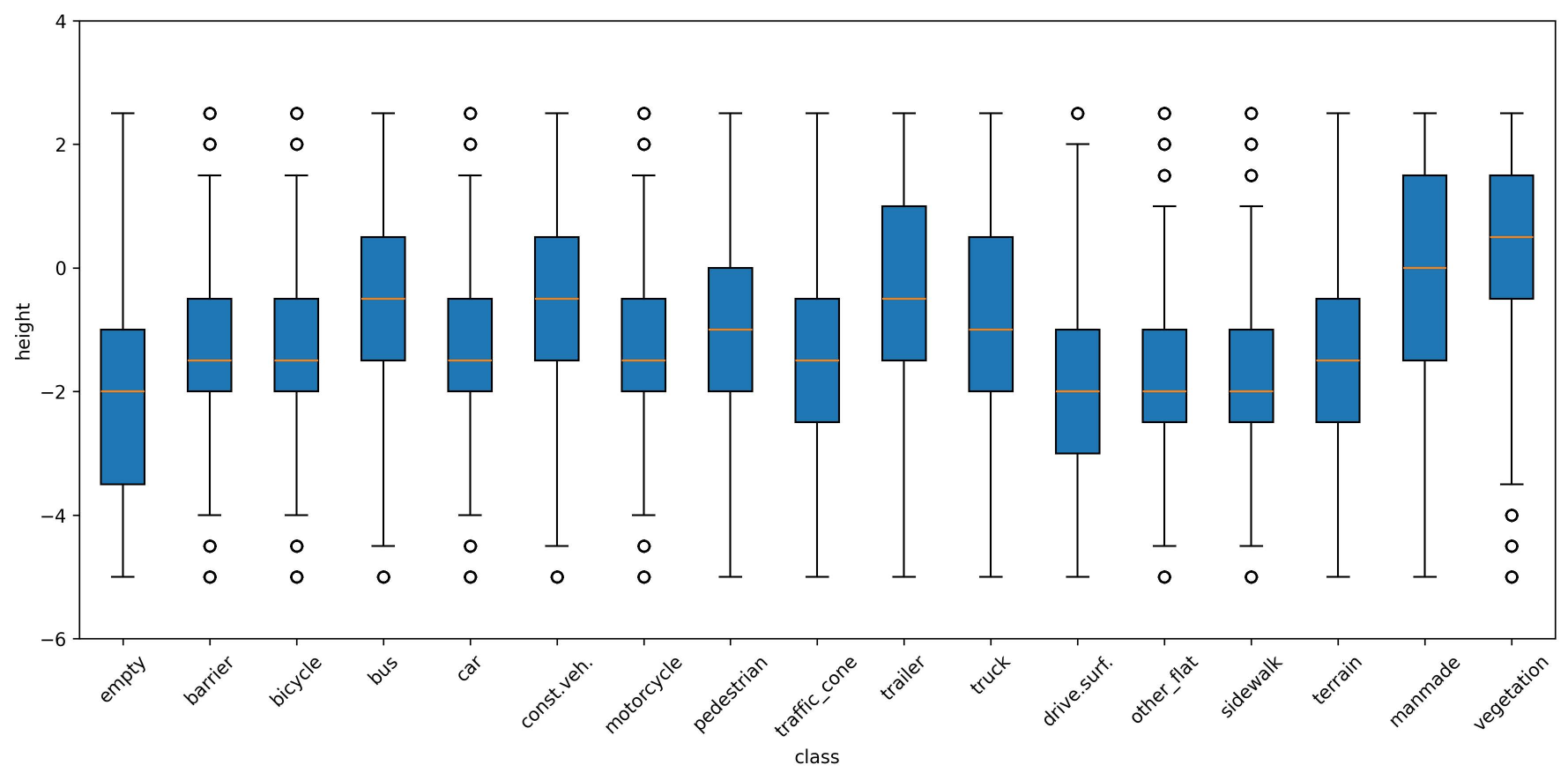}
\vspace{-2.0em}
\end{center}
   \caption{
   Height distribution of different classes on nuScenes-SurroundOcc.}\label{nuScenes-SurroundOcc}
\vspace{-1.5em}
\end{figure}

In Figure \ref{nuScenes-SurroundOcc}, the nuScenes-SurroundOcc dataset \cite{SurroundOcc} exhibits marked variations in the height distribution of different object categories. Motivated by this observation, we propose SliceSemOcc, a vertical-slice-based feature representation and fusion framework. Specifically, we first sample voxel features along the height-axis at two scales. Global slices span the entire height range, while local slices focus on key height intervals within a limited band. We then integrate global and local slice features through a bidirectional cross attention mechanism, enabling joint modeling of fine details and broader context. To overcome the uniform weighting of all height layers by conventional channel attention, we propose the SEAttention3D module. This module preserves resolution along the height-axis during average pooling and dynamically assigns channel attention weights to each height layer. These innovations enable SliceSemOcc to significantly enhance the model’s ability to represent diverse semantic structures in complex 3D scenes.

Our contributions are summarized as follows: 
\begin{itemize}
  \item We develop the Vertical Slice Fusion module, which employs a dual scale vertical slice strategy by extracting global slices across the full height range and local slices focused on key height bands. These representations are then fused using bidirectional cross attention to capture both overall structure and fine detail.
  \item We propose the SEAttention3D channel attention mechanism, which preserves height-axis resolution during pooling and dynamically assigns distinct channel weights to each height layer, thereby significantly enhancing the representation of vertical structures.
  \item We perform extensive experiments on the nuScenes-SurroundOcc and nuScenes-OpenOccupancy datasets, demonstrating that SliceSemOcc delivers substantial mIoU gains. We also conduct ablation experiments to validate the effectiveness of SliceSemOcc.
\end{itemize}

\section{Related Work}
\label{sec:related}

\subsection{Camera-only-based Environment Perception}

In recent years, the low cost and versatility of surround-view cameras have driven widespread interest in camera based perception algorithms for autonomous driving. Contemporary BEV methods transform feature representations from image space into BEV space. One approach follows the lifting paradigm introduced by LSS \cite{lss}, in which a depth map is explicitly predicted and used to project multi-view image features onto the BEV plane \cite{BEVDet,Bevdepth}. A second approach adopts the query-based framework of DETR3D \cite{Detr3d}, employing learnable queries and cross attention to extract information directly from image features \cite{BEVFormer,PolarFormer}. Although both approaches can provide explicit or implicit depth cues, monocular cameras are well known to struggle with precise depth estimation, often capturing only relative depth relationships rather than absolute distances. Consequently, more reliable depth references often require model integration of LiDAR data to enhance monocular depth estimation or LiDAR supervision as in the BEVDet series. While these techniques successfully compress information into the BEV plane, they may discard some of the scene’s inherent 3D structure. Voxel based representations overcome this limitation by preserving a holistic view of 3D space and are therefore well suited to tasks such as 3D semantic segmentation and full view segmentation.

\subsection{LiDAR-only-based Environment Perception}

LiDAR excels at high-precision depth measurement and therefore offers distinct advantages for capturing object geometry and 3D position. Representative works \cite{cylinder3d,DRINet++} extract point-cloud features using dense or sparse 3D convolutions to achieve high-precision object detection. At the same time researchers have applied LiDAR point-cloud features to 3D semantic occupancy prediction \cite{4d-occ,occ4cast}. However uneven point-cloud density generated by LiDAR can significantly impair perception performance and its inherent lack of semantic information often leads to bias in object category recognition. It is therefore necessary to introduce supplementary information to provide more comprehensive semantic guidance.

\subsection{Camera-LiDAR fusion-based environment perception}

Given the inherent limitations of individual sensors, recent work has increasingly turned to multi‐sensor fusion techniques \cite{f-pointnet,Ipod,ContFuse} to address these weaknesses and improve overall scene understanding. Representative approaches such as BEVFusion \cite{BEVFusion,Bevfusion2} encode features from LiDAR point clouds and surround‐view cameras into a unified BEV representation. This strategy reduces false positives and negatives caused by LiDAR reflections in rain or fog and addresses the depth estimation shortcomings of monocular cameras, enabling more accurate long-range detection. Multimodal semantic occupancy prediction methods such as OccFusion [25] and OpenOccupancy [35] project camera and LiDAR features into a unified voxel space. They then fuse these features using channel or spatial attention for occupancy inference. These approaches all benefit from multi source data to enhance detection and occupancy accuracy but concentrate fusion at the planar or global voxel level. They fail to exploit hierarchical representations or perform differentiated fusion across height layers, limiting their ability to capture vertical semantic complementarity.

\section{Method}
\label{sec:method}
\subsection{Problem Statement}

Our goal is to predict the 3D occupancy of the surrounding scene given LiDAR sweeps $L$ and surround‐view images $I = \{I^1, I^2, \ldots, I^N\}$ from $N$ cameras. Accordingly, the problem can be formulated as:

\begin{equation}
  V=F\left(I^1, I^2, \cdots I^N, L \right) \label{Eq.1}
  \end{equation}

Here, $F$ denotes the multi‐sensor fusion framework for 3D spatial occupancy prediction. The resulting 3D occupancy volume $V$ assigns each grid cell a semantic label in [0, 16], where 0 indicates an empty cell. In our formulation, a label of 0 corresponds to an empty cell.

\subsection{Overall Architecture}
\begin{figure*}[t]
\begin{center}
\includegraphics[width=\textwidth,trim=2mm 2mm 2mm 2mm,
  clip]{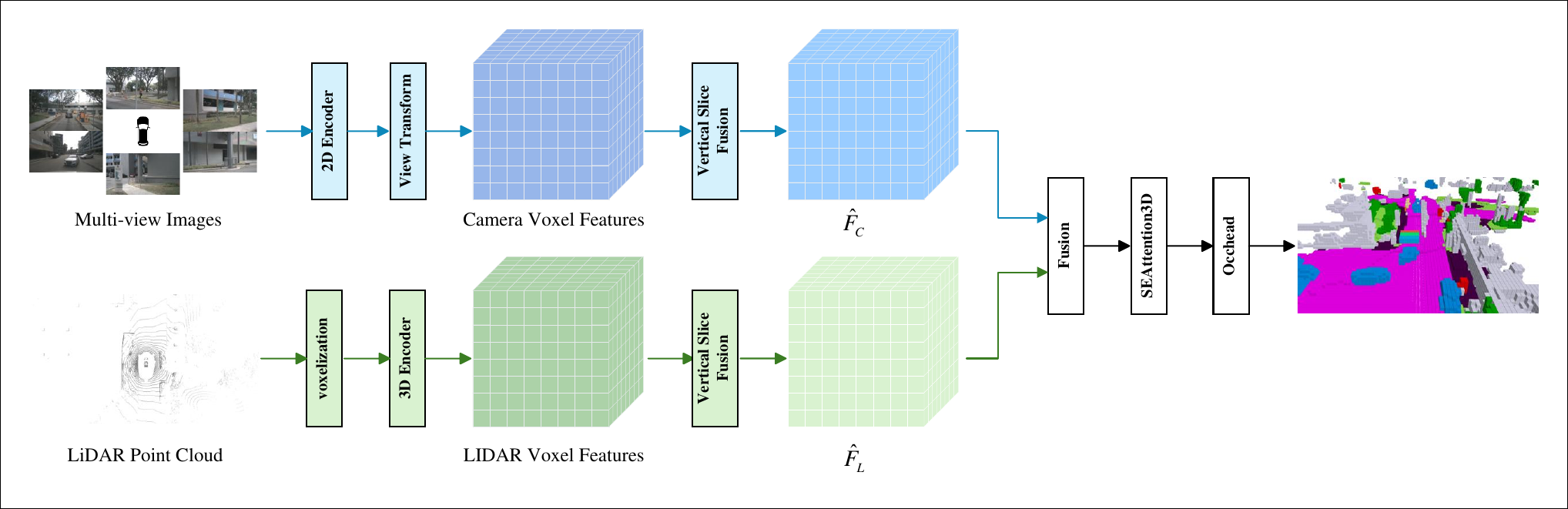}
\vspace{-1.3em}
\end{center}
   \caption{Overall architecture of SliceSemOcc.}
\label{SliceSemOcc}
\vspace{-1.0em}
\end{figure*}

We propose the SliceSemOcc framework (Fig. \ref{SliceSemOcc}), which takes a dense LiDAR point cloud and surround‐view images as input. On the camera branch, a 2D backbone extracts multi‐view image features, which are then projected into the 3D voxel space via a view–voxel transformation module, yielding visual voxel features$F_C \in \mathbb{R}^{B\times C\times X\times Y\times Z}$.Concurrently, on the LiDAR branch, the raw point cloud is voxelized and processed by a 3D backbone to produce point‐cloud voxel features$F_L \in \mathbb{R}^{B\times C\times X\times Y\times Z}.$Both $F_C$ and $F_L$ pass through the Vertical Slice Fusion module which divides each along the height-axis into global slices covering the entire range and local slices focusing on key height bands. Each slice is first processed by our SEAttention3D module. Then, a bidirectional cross attention mechanism fuses global and local slice features through mutual refinement, producing refined features$\hat F_C, \hat F_L \in \mathbb{R}^{B\times C\times X\times Y\times Z}$. Finally, these refined visual and LiDAR voxel features are concatenated and passed through a 3D convolutional decoding head to generate the final occupancy probability volume and semantic label predictions.

\subsection{SEAttention3D Module}
\begin{figure*}[t]
\begin{center}
\includegraphics[width=0.4\textwidth,trim=2mm 2mm 2mm 2mm,
  clip]{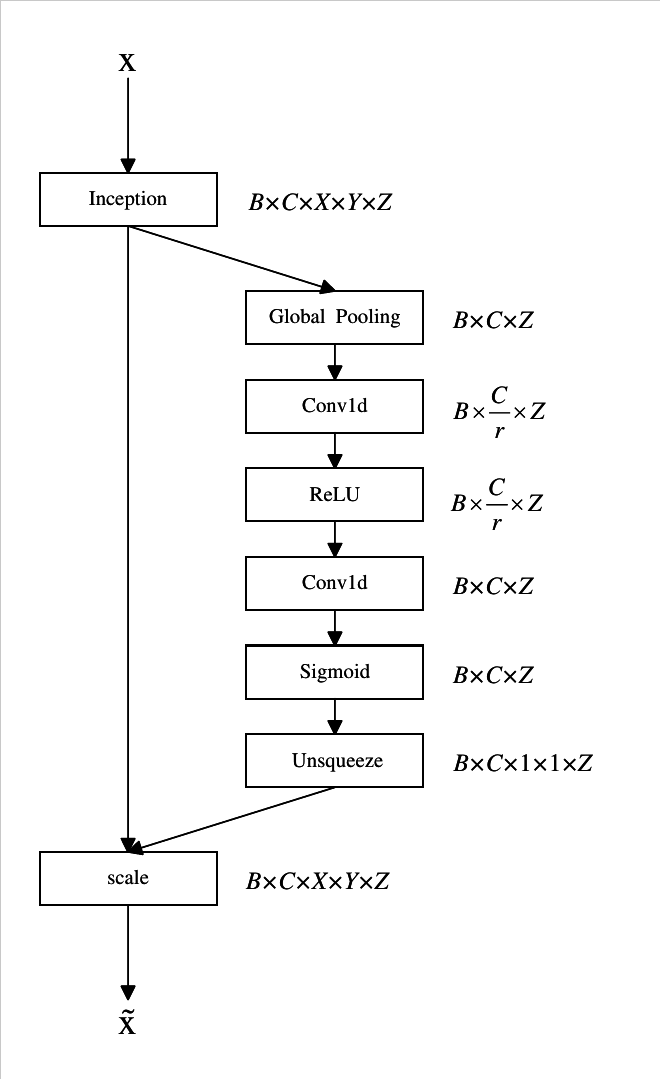}
\vspace{-1.3em}
\end{center}
   \caption{Detailed architecture of the SEAttention3D module.}
\label{SEAttention3D}
\vspace{-1.0em}
\end{figure*}

Conventional channel‐attention mechanisms (e.g., SENet) apply global average pooling over all spatial dimensions, causing every height layer to share the same channel weights and thus failing to discriminate the importance of features at different heights. To overcome this limitation, we design the SEAttention3D module (see Fig. \ref{SEAttention3D}), whose core idea is to preserve resolution along the height-axis during pooling. The detailed workflow is as follows.

\paragraph{Squeeze}

Given input voxel features $\mathcal{X}\in\mathbb{R}^{B\times C\times X\times Y\times Z}$, we perform global average pooling exclusively over the spatial plane dimensions $X$ and $Y$:

\begin{equation}
  S_{b,c,z}
  =\frac{1}{XY}\sum_{i=1}^{X}\sum_{j=1}^{Y}
  \mathcal{X}_{b,c,i,j,z},
  \label{Eq.3}
  \end{equation}

This produces a 3D tensor $S \in \mathbb{R}^{B \times C \times Z}$. 

\paragraph{Excitation}
We treat $S$ as a sequence and apply two 1D convolutional layers per channel: the first layer reduces the channel dimension from $C$ to $\tfrac{C}{r}$ (where $r$ is a predefined reduction ratio) and is followed by a ReLU activation; the second layer restores the channel dimension from $\tfrac{C}{r}$ back to $C$ and is followed by a Sigmoid. This yields an output tensor $S'$ of the same shape $[B, C, Z]$. We then expand $S'$ along the spatial dimensions:

\begin{equation}
  \widetilde{\mathcal{S}}_{b,c,i,j,z} = \mathrm{reshape}(S'_{b,c,z})\;\in\;\mathbb R^{B\times C\times 1\times 1\times Z} \label{Eq.3}
\end{equation}

We then multiply elementwise by the original input $\mathcal{X}$, completing the dynamic weighting of each channel at different height levels.

\begin{equation}
  \mathcal{Y}_{b,c,i,j,z} = \mathcal{X}_{b,c,i,j,z}\times \widetilde{\mathcal{S}}_{b,c,i,j,z} \label{Eq.3}
  \end{equation}

With this design SEAttention3D assigns differentiated channel responses at each height level which strengthens semantic information in critical height ranges while suppressing noise in non critical layers. This module endows the network with finer, more hierarchically structured representations in the 3D voxel space. 

\subsection{Vertical Slice Fusion Module}
To fully exploit the semantic information inherent in different height layers of the voxel space, we design the vertical slice fusion module VSF, which comprises two parts global slices and local slices and a cross attention fusion mechanism that integrates features from both.

\begin{figure*}[t]
\begin{center}
\includegraphics[width=1.0\textwidth,trim=2mm 2mm 2mm 2mm,
  clip]{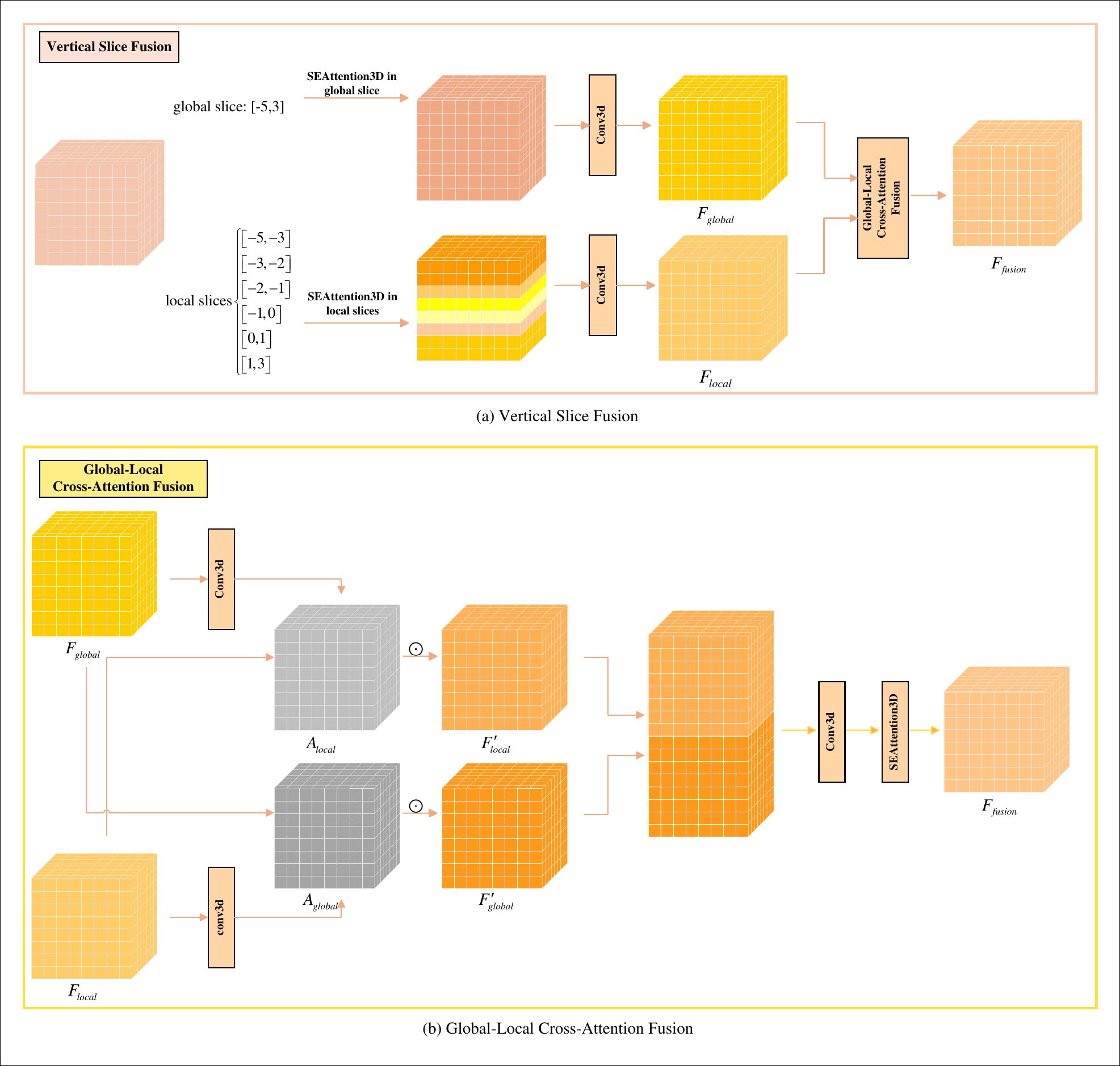}
\vspace{-1.3em}
\end{center}
   \caption{Detailed architecture of (a) Vertical Slice Fusion (VSF) module and (b) Global–Local  cross Attention Fusion module}
\label{VSF}
\vspace{-1.0em}
\end{figure*}

\paragraph{Global and Local Slices}
As illustrated in Fig. \ref{nuScenes-SurroundOcc}, the nuScenes SurroundOcc dataset reveals significant differences in object height distributions across categories. Small objects such as pedestrians and traffic barriers are primarily concentrated within the $[-2,2]$ m range, whereas large objects such as trucks and buildings span a much broader height extent. To more effectively sample local slices that reflect small‐object height distributions, we partition the full height range $[-5,3]$ m into six subintervals: $[-5,-3]$, $[-3,-2]$, $[-2,-1]$, $[-1,0]$, $[0,1]$, and $[1,3]$, each serving as the basis for a local slice. Simultaneously, we treat the entire span $[-5,3]$ m as a single interval to generate a global slice. As illustrated in Fig. \ref{VSF}, for each global and local slice, we independently apply the SEAttention3D module to extract channel‐weighted voxel features. We then integrate these features via a 3D convolutional module to produce the global voxel feature $F_{\text{global}}$ and the local voxel feature $F_{\text{local}}$, respectively.

\paragraph{Global-Local Cross‑Attention Fusion}
Global slice attention excels at modeling long‐range dependencies and holistic structure but tends to overlook fine details of nearby small objects. Conversely, local slice attention captures detailed contours of small‐scale targets but may lack broader scene context. To address these complementary shortcomings, we integrate a bidirectional cross‐attention mechanism within the VSF module (Fig. \ref{VSF}), enabling global attention to guide local features toward critical regions while local attention enriches global features with fine‐grained details.

Specifically, we apply a 3D convolution to both the global voxel feature $F_{\text{global}}$ and the local voxel feature $F_{\text{local}}$ to reduce their channels from $C$ to one. This operation independently generates the attention maps $A_{\text{global}}$ and $A_{\text{local}}$. We then weight the local feature by the global attention and the global feature by the local attention. For detailed descriptions of the Global Local Cross Attention Fusion module refer to Fig. \ref{VSF}.

\begin{equation}
  \begin{aligned}
    F'_\text{global} \;=\; A_\text{local}\;\odot\;F_\text{global} \\[6pt]
    F'_\text{local} \;=\; A_\text{global}\;\odot\;F_\text{local}
  \end{aligned}
\end{equation}


Where $\odot$ denotes element-wise multiplication, and $F'_{\mathrm{global}}$ and $F'_{\mathrm{local}}$ refer to the  cross attention-calibrated global and local features, respectively.

We first concatenate $F'_{\mathrm{global}}$ and $F'_{\mathrm{local}}$ along the channel axis to produce a fused feature of dimension $2C$. This fused tensor is then passed through a 3D convolutional layer to reduce its channel dimension back to $C$. Finally, we re-apply the SEAttention3D module to perform fine-grained weighting along the height-axis, yielding the final fused voxel feature $F_{\mathrm{fusion}}$.

\subsection{Loss}

To train our proposed baseline, we optimize the network using the focal loss \cite{Focal-Loss}, LovaszSoftmax loss \cite{Lovasz-Softmax} .Following \cite{MonoScene} , we also utilize affinity loss $\mathcal{L}_{\mathrm{scal}}^{\mathrm{geo}}$
and $\mathcal{L}_{\mathrm{scal}}^{\mathrm{sem}}$ to optimize the scene-wise and class-wise metrics.


\begin{equation}
  \mathcal{L}_{\text {total }}=\mathcal{L}_{\mathrm{focal}}+\mathcal{L}_{\mathrm{lovasz}}+\mathcal{L}_{\mathrm{scal}}^{\mathrm{geo}}+\mathcal{L}_{\mathrm{scal}}^{\mathrm{sem}} \label{Eq.7}
\end{equation}

\section{Experiments}
\label{sec:experiment}

\subsection{Implementation Details}

In the camera branch, we employ ResNet-50 or ResNet-101 \cite{Deep-Residual-Learning} equipped with an FPN \cite{Feature-Pyramid-Networks} on both datasets. The view transformer produces 3D feature volumes. in our implementation, we adopt the projection method proposed in \cite{InverseMatrixVT3D} to aggregate visual features. In the LiDAR branch we voxelize ten LiDAR sweeps and use a voxel encoding scheme developed for the nuScenes dataset. To extract 3D feature volumes, we use VoxelNet \cite{VoxelNet} as the 3D backbone.
We evaluate our method on the challenging nuScenes dataset \cite{nuScenes}, using ground-truth occupancy labels provided by OpenOccupancy \cite{OpenOccupancy} and SurroundOcc \cite{SurroundOcc}. For SurroundOcc, the annotation covers an $X Y$ range of –50 m to 50 m and a $Z$ range of –5 m to 3 m, resulting in an occupancy grid resolution of $200 \times 200 \times 16$. For OpenOccupancy, the evaluation ranges are set to [–51.2 m, 51.2 m] in $X$ and $Y$, and [–5 m, 3 m] in $Z$, yielding a grid of $40 \times 512 \times 512$.

\begin{table*}[t]
	\setlength{\tabcolsep}{0.0035\linewidth}
	\newcommand{\classfreq}[1]{{~\tiny(\semkitfreq{#1}\%)}}  %
	\centering
    \caption{\textbf{3D semantic occupancy prediction results on nuScenes validation set}. The C and L denotes camera and LiDAR, respectively.}

   \resizebox{1\linewidth}{!}{
	\begin{tabular}{l|c |c | c | c c c c c c c c c c c c c c c c}

		\toprule
		Method
		& \makecell[c]{Backbone}
		& \makecell[c]{Input Modality}
		
            & \makecell[c]{mIoU}
		& \rotatebox{90}{\textcolor{barrier}{$\blacksquare$} barrier} 
		& \rotatebox{90}{\textcolor{bicycle}{$\blacksquare$} bicycle}
		& \rotatebox{90}{\textcolor{bus}{$\blacksquare$} bus} 
		& \rotatebox{90}{\textcolor{car}{$\blacksquare$} car} 
		& \rotatebox{90}{\textcolor{const. veh.}{$\blacksquare$} const. veh.} 
		& \rotatebox{90}{\textcolor{motorcycle}{$\blacksquare$} motorcycle} 
		& \rotatebox{90}{\textcolor{pedestrian}{$\blacksquare$} pedestrian} 
		& \rotatebox{90}{\textcolor{traffic cone}{$\blacksquare$} traffic cone} 
		& \rotatebox{90}{\textcolor{trailer}{$\blacksquare$} trailer} 
		& \rotatebox{90}{\textcolor{truck}{$\blacksquare$} truck} 
		& \rotatebox{90}{\textcolor{drive. suf.}{$\blacksquare$} drive. suf.} 
		& \rotatebox{90}{\textcolor{other flat}{$\blacksquare$} other flat} 
		& \rotatebox{90}{\textcolor{sidewalk}{$\blacksquare$} sidewalk} 
		& \rotatebox{90}{\textcolor{terrain}{$\blacksquare$} terrain} 
		& \rotatebox{90}{\textcolor{manmade}{$\blacksquare$} manmade} 
		& \rotatebox{90}{\textcolor{vegetation}{$\blacksquare$} vegetation} \\
		\midrule
		MonoScene \cite{MonoScene} & R101-DCN & C  & 7.3 & 4.0  & 0.4  &  8.0 &  8.0 & 2.9  & 0.3  &  1.2& 0.7& 4.0 & 4.4 & 27.7 & 5.2  & 15.1& 11.3 & 9.0 & 14.9\\

    BEVFormer \cite{BEVFormer} & R101-DCN & C & 16.7 &14.2& 6.5 &23.4& 28.2& 8.6& 10.7 &6.4& 4.0 &11.2& 17.7 &37.2& 18.0& 22.8& 22.1 &13.8& 22.2 \\

    TPVFormer \cite{TPVFormer} & R101-DCN & C & 17.10  & 15.96 & 5.31 & 23.86 & 27.32 & 9.79 & 8.74 & 7.09 & 5.20 & 10.97 & 19.22 & 38.87 & 21.25 & 24.26 & 23.15 & 11.73 & 20.81 \\
    
    C-CONet \cite{OpenOccupancy} & R101 & C & 18.4 &18.6 &10.0 &26.4 &27.4& 8.6 &15.7 &13.3 &9.7 &10.9 &20.2& 33.0& 20.7& 21.4& 21.8& 14.7 &21.3 \\

    OccFormer  & R101 & C  & 20.1 &21.1& 11.3 &28.2& 30.3& 10.6& 15.7 &14.4& 11.2 &14.0& 22.6 &37.3&22.4 & 24.9& 23.5 &15.2& 21.1 \\

    RenderOcc \cite{RenderOcc} & R101 & C & 19.0 & 19.7 & 11.2 & 28.1 & 28.2 & 9.8 & 14.7 & 11.8 & 11.9 & 13.1 & 20.1 & 33.2 & 21.3 & 22.6 & 22.3 & 15.3 & 20.9 \\

    SurroundOcc \cite{SurroundOcc} & R101-DCN & C & 20.3 &20.5 &11.6 &28.1 &30.8& 10.7 &15.1 &14.0 &12.0 &14.3 &22.2& 37.2& 23.7& 24.4& 22.7& 14.8 &21.8 \\

    LMSCNet \cite{Lmscnet} & - & L & 14.9 & 13.1&  4.5 & 14.7  & 22.1  & 12.6  &  4.2 & 7.2 & 7.1&  12.2&  11.5& 26.3 & 14.3  & 21.1 & 15.2  & 18.5 & 34.2 \\
    L-CONet \cite{OpenOccupancy} & - & L & 17.7 & 19.2 & 4.0& 15.1 & 26.9 & 6.2 & 3.8 &  6.8& 6.0& 14.1&  13.1& 39.7 & 19.1 & 24.0  & 23.9 & 25.1 & 35.7 \\

    M-CONet \cite{OpenOccupancy} & - & C+L & 24.7  & 24.8 & 13.0& 31.6 & 34.8 & 14.6 & 18.0  &  20.0& 14.7& 20.0&  26.6& 39.2 & 22.8 & 26.1  & 26.0 & 26.0 & 37.1 \\
  
    Co-Occ \cite{Co-Occ} &R101 & C\&L &27.1  &28.1 &16.1 &34.0 &37.2 &17.0&21.6 &20.8 &15.9 &21.9 &28.7&\textbf{42.3}&\textbf{25.4} &\textbf{29.1} &\textbf{28.6} &28.2 &38.0  \\

    OccFusion \cite{OccFusion} & R101-DCN & C+L &  27.6& 25.2& \textbf{19.9}& 34.78& 36.2& \textbf{20.0}& 23.1& 25.3& 17.5& 22.7& 30.1& 39.5& 23.23& 25.7& 27.6 &29.5 &40.6 \\

    \hline
    Slice-Occ (Ours)& R101-DCN   & C\&L     & \textbf{28.2}      & \textbf{30.1}  & \textbf{19.9}  & \textbf{36.5}  & \textbf{38.9}  & 19.5  & \textbf{24.7}  & \textbf{26.0}  & \textbf{17.7}  & \textbf{23.9}  & \textbf{32.2}  & 37.1  & 22.0 & 26.5  & 25.3  & \textbf{30.4}  & \textbf{40.8} \\
		\bottomrule
	\end{tabular}}
	\label{SurroundOcc}
\end{table*}
\begin{table*}[t]
	\setlength{\tabcolsep}{0.0035\linewidth}
	\newcommand{\classfreq}[1]{{~\tiny(\semkitfreq{#1}\%)}}  %
	\centering
    \caption{\textbf{3D semantic occupancy prediction results on nuScenes-Occupancy validation set}. The C and L denotes camera and LiDAR, respectively. }

   \resizebox{1\linewidth}{!}{
	\begin{tabular}{l|c | c | c c c c c c c c c c c c c c c c}

		\toprule
		Method
		& \makecell[c]{Backbone}

            & \makecell[c]{mIoU}
		& \rotatebox{90}{\textcolor{barrier}{$\blacksquare$} barrier} 
		& \rotatebox{90}{\textcolor{bicycle}{$\blacksquare$} bicycle}
		& \rotatebox{90}{\textcolor{bus}{$\blacksquare$} bus} 
		& \rotatebox{90}{\textcolor{car}{$\blacksquare$} car} 
		& \rotatebox{90}{\textcolor{const. veh.}{$\blacksquare$} const. veh.} 
		& \rotatebox{90}{\textcolor{motorcycle}{$\blacksquare$} motorcycle} 
		& \rotatebox{90}{\textcolor{pedestrian}{$\blacksquare$} pedestrian} 
		& \rotatebox{90}{\textcolor{traffic cone}{$\blacksquare$} traffic cone} 
		& \rotatebox{90}{\textcolor{trailer}{$\blacksquare$} trailer} 
		& \rotatebox{90}{\textcolor{truck}{$\blacksquare$} truck} 
		& \rotatebox{90}{\textcolor{drive. suf.}{$\blacksquare$} drive. suf.} 
		& \rotatebox{90}{\textcolor{other flat}{$\blacksquare$} other flat} 
		& \rotatebox{90}{\textcolor{sidewalk}{$\blacksquare$} sidewalk} 
		& \rotatebox{90}{\textcolor{terrain}{$\blacksquare$} terrain} 
		& \rotatebox{90}{\textcolor{manmade}{$\blacksquare$} manmade} 
		& \rotatebox{90}{\textcolor{vegetation}{$\blacksquare$} vegetation} \\
		\midrule
    MonoScene \cite{MonoScene}  & C & 6.9 & 7.1 & 3.9 & 9.3 & 7.2 & 5.6 & 3.0 & 5.9 & 4.4 & 4.9 & 4.2 & 14.9 & 6.3 & 7.9 & 7.4 & 10.0 & 7.6 \\
    
    TPVFormer \cite{TPVFormer}  & C & 7.8 & 9.3 & 4.1 & 11.3 & 10.1 & 5.2 & 4.3 & 5.9 & 5.3 & 6.8 & 6.5 & 13.6 & 9.0 & 8.3 & 8.0 & 9.2 & 8.2 \\
    
    LMSCNet \cite{Lmscnet} & L  & 11.5 & 12.4 & 4.2 & 12.8 & 12.1 & 6.2 & 4.7 & 6.2 & 6.3 & 8.8 & 7.2 & 24.2 & 12.3 & 16.6 & 14.1 & 13.9 & 22.2 \\

    JS3C-Net \cite{JS3C-Net} & L  & 12.5 & 14.2 & 3.4 & 13.6 & 12.0 & 7.2 & 4.3 & 7.3 & 6.8 & 9.2 & 9.1 & 27.9 & 15.3 & 14.9 & 16.2 & 14.0 & \textbf{24.9} \\
    3DSketch \cite{3dsketch} &C\&D &10.7 & 12.0 & 5.1 & 10.7 & 12.4 & 6.5 & 4.0 & 5.0 & 6.3 & 8.0 & 7.2 & 21.8 & 14.8 & 13.0 & 11.8 & 12.0 & 21.2\\
    AICNet \cite{Anisotropic} & C\&D & 10.6 & 11.5 & 4.0 & 11.8 & 12.3 & 5.1 & 3.8 & 6.2 & 6.0 & 8.2 & 7.5 & 24.1 & 13.0 & 12.8 & 11.5 & 11.6 & 20.2 \\

    M-CONet \cite{OpenOccupancy}  & C\&L  & 20.1 & 23.3 & 13.3 & 21.2 & 24.3 & 15.3 & 15.9 & 18.0 & 13.3 & 15.3 & 20.7 & 33.2 & 21.0 & 22.5 & 21.5 & 19.6 & 23.2 \\

    Co-Occ \cite{Co-Occ} & C\&L  & 21.9 & 26.5 & 16.8 & 22.3 & 27.0 & 10.1 & 20.9 & 20.7 & 14.5 & 16.4 & 21.6 & 36.9 & \textbf{23.5} & \textbf{25.5} & 23.7 & 20.5 & 23.5 \\

    OccLoff \cite{OccLoff} & C\&L  &\textbf{22.9} &26.7 &17.2 &22.6 &26.9 &\textbf{16.4} &22.6 &24.7 &16.4 &16.3 &22.0 &\textbf{37.5} &22.3 &25.3 &\textbf{23.9} &21.4 &24.2\\
    \hline
    SliceSemOcc  & C\&L & \textbf{22.9} & \textbf{27.4}  & \textbf{20.5}  & \textbf{23.7}  & \textbf{28.6}  & 14.5  &\textbf{24.8}  & \textbf{27.5}  & \textbf{15.2}  & \textbf{19.1}  & \textbf{23.9}  & 31.8  & 20.4  & 22.2  & 20.0  & \textbf{22.3 } & 24.0 \\
		\bottomrule
	\end{tabular}}
	\label{OpenOccupancy}
\end{table*}

\begin{figure*}[t]
\begin{center}
\includegraphics[width=\textwidth,trim=2mm 2mm 2mm 2mm,
  clip]{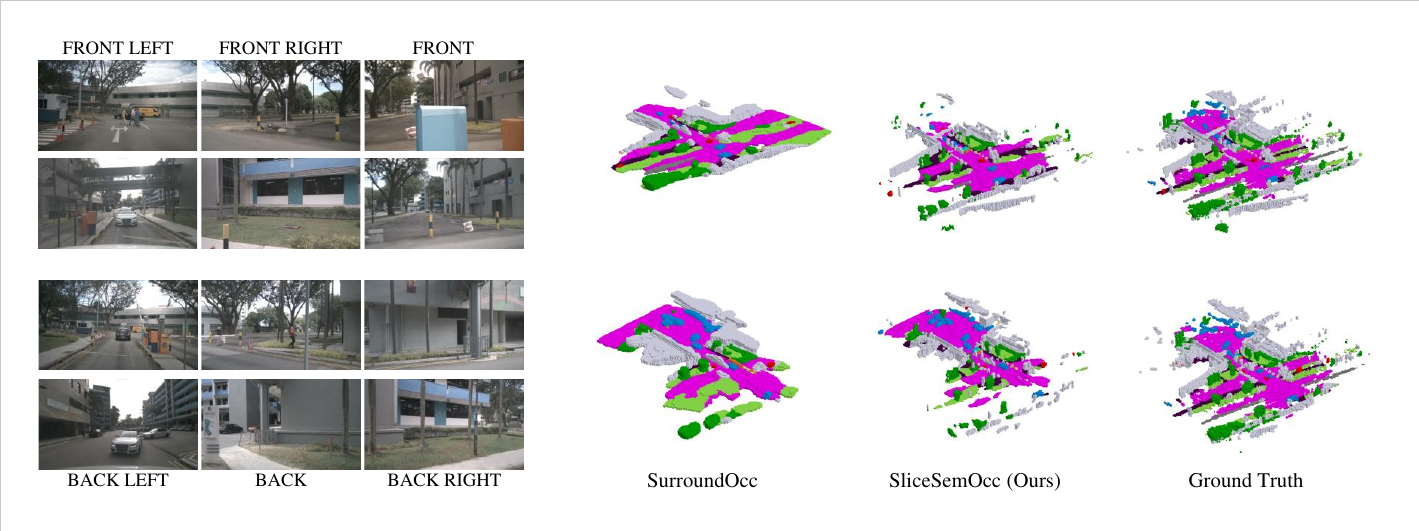}
\vspace{-1.3em}
\end{center}
   \caption{Qualitative performance on nuScenes-SurroundOcc dataset.}
\label{visual2}
\vspace{-1.0em}
\end{figure*}

\subsection{Main Results}
To comprehensively assess the performance of the proposed SliceSemOcc framework on 3D semantic occupancy prediction, we evaluate it on the validation sets of two public datasets, nuScenes-SurroundOcc and nuScenes-OpenOccupancy. The final results, shown in Tables \ref{SurroundOcc} and \ref{OpenOccupancy}, indicate that SliceSemOcc achieves overall mIoU scores of 28.2\% on nuScenes-SurroundOcc and 22.9\% on nuScenes-OpenOccupancy, confirming the synergistic benefits of our vertical-slice fusion and SEAttention3D modules. In Fig. \ref{visual2}, we provide visualization results on nuScenes-SurroundOcc dataset.

On nuScenes-SurroundOcc dataset, SliceSemOcc raises the overall mIoU from 24.7\% (M-CONet) to 28.2\%, a relative gain of 14.2\%. Gains are especially pronounced for small objects—barrier (+21.3\%), bicycle (+53.1\%), motorcycle (+37.2\%), pedestrian (+30.0\%), and traffic cone (+20.4\%). Solid improvements are also seen for medium-to-large objects (car, bus, truck, etc.). This demonstrates that local slices densely sample small-object height ranges to boost fine-scale discrimination, while global slices and  cross attention capture macro-structures, together far surpassing M-CONet.

On nuScenes-OpenOccupancy dataset, SliceSemOcc improves mIoU from 20.1\% (M-CONet) to 22.9\%, a relative uplift of 13.9\%. Compared to M-CONet, SliceSemOcc delivers substantial gains on small-object categories and balanced improvements on medium/large classes and overall structure, fully validating the complementary efficacy of our vertical-slice fusion strategy and SEAttention3D module in enhancing multi-modal 3D semantic occupancy prediction’s accuracy and robustness.

Notably, SliceSemOcc exhibits lower IoU on the driveable surface, other flat, sidewalk, and terrain classes. These four categories all correspond to “flat” ground surfaces—roadway, flat pavement, sidewalks, and natural terrain—that share very similar geometric profiles and reflection characteristics (LiDAR intensity, RGB reflectance). Although our method allocates channel attention across height layers, slices at adjacent heights with nearly identical features still receive similar weights, making it challenging to effectively distinguish these flat-ground categories.

\subsection{Ablation Study}
\paragraph{Global and Local Slices}
Our method employs both global and local vertical slices to construct voxel features: the global slice spans the entire height range, while the local slices concentrate on key height bands. To quantify their individual contributions, we conducted an ablation study. As shown in Table \ref{Global_and_Local_Slices}, both the global and local slicing strategies independently yield performance gains.

\begin{table}[htbp]
    \centering
    \caption{Ablation study of Global and Local Slices.}

    \begin{tabular}{cc|c|c}
    \toprule
    Local & Global & mIoU & Memory  \\    
    \midrule
    \XSolidBrush & \XSolidBrush &    26.98 &    5.6GB \\
    \Checkmark & \XSolidBrush &  27.24 &  5.7GB \\
    \XSolidBrush & \Checkmark &   27.41&   5.7GB\\
    \Checkmark & \Checkmark &   28.21&   6.2GB\\
    \bottomrule
  \end{tabular}
  \label{Global_and_Local_Slices}
\end{table}

\begin{table}[htbp]
    \centering
    \caption{Ablation study results on the local slice sampling strategy used in the Vertical Slice Fusion moudle.}

    \begin{tabular}{c|c}
    \toprule
    Height Ranges & mIoU\\
    \midrule
    \mbox{[-5, -4],[-4, -3],[-3, -2],[-2, -1],[-1, 0],[0, 1],[1, 2],[2, 3]} & 27.67  \\
    \mbox{[-5, -3],[-3, -1],[-1, 1],[1, 3]} &  27.75 \\
    \mbox{[-5, -3],[-3, -2],[-2, -1],[-1, 0],[0, 1],[1, 3]} &  28.21 \\
    \bottomrule
  \end{tabular}
  \label{Local_Slice_Sampling_Strategy}
\end{table}



\paragraph{Local Slice Sampling Strategy}

The local slice sampling strategy in this paper is driven by the object height distribution found in the nuScenes-SurroundOcc dataset. To quantify its impact, we performed an ablation study comparing it against a uniform sampling baseline. As shown in Table \ref{Local_Slice_Sampling_Strategy}, sampling local slices according to the nuScenes-SurroundOcc height distribution yields the best performance.

\paragraph{SEAttention3D Module}
We integrate the SEAttention3D module into our framework. To quantify its impact, we conduct an ablation study against the vanilla SENet channel‐attention mechanism. When using the standard SENet, the model achieves an mIoU of 28.08\%. By replacing it with SEAttention3D, the mIoU increases to 28.21\%, demonstrating that SEAttention3D contributes a measurable performance improvement.

\paragraph{Fusion Strategy}
We incorporate the Global–Local Cross-Attention Fusion module in our framework. To assess its impact, we conduct an ablation study: when fusing global and local voxel features using only a simple concatenation operation, the model achieves an mIoU of 27.97\%. By contrast, employing the Global–Local Cross-Attention Fusion module to merge these features raises the mIoU to 28.21\%, demonstrating its effectiveness.

\paragraph{Computational Performance of Our Model}
Table \ref{Global_and_Local_Slices} reports the computational performance of SliceSemOcc during validation. Although the inclusion of the VSF module incurs an additional 0.6 GB of GPU memory, it delivers a 1.14\% increase in mIoU, with particularly pronounced gains on most small‐object categories. We therefore consider this modest memory overhead to be well justified.

\section{Conclusion}
In this work, we present a novel approach named SliceSemOcc for 3D semantic occupancy prediction. SliceSemOcc framework extracts voxel features along the height-axis using both global and local vertical slices that capture semantic information at different height levels. To enhance sampling, SliceSemOcc draws local slices according to the object height distribution in nuScenes-SurroundOcc dataset and integrate global and local features with a Global Local Cross Attention Fusion module. We further propose the SEAttention3D module so that the network can assign distinct channel attention weights to each vertical slice. Experimental results demonstrate the effectiveness of SliceSemOcc framework and confirm the importance of height information in occupancy prediction. In future work, we will strive to make SliceSemOcc as lightweight as possible to reduce the computational burden of training and validation.

\bibliographystyle{splncs04}
\bibliography{main}
\end{document}